%% file: main.tex
\documentclass[letterpaper, 10 pt, conference]{ieeeconf}  

\IEEEoverridecommandlockouts                              
\usepackage[utf8]{inputenc}
\usepackage[T1]{fontenc}
\usepackage{graphicx}
\usepackage{pifont}
\usepackage{multirow}

\usepackage{amsmath} 

\usepackage[table,xcdraw]{xcolor}
\usepackage{amssymb}

\definecolor{rblue}{rgb}{0,0.5,1}
\definecolor{awesome}{rgb}{1.0, 0.13, 0.32}
\definecolor{hollywoodcerise}{rgb}{0.96, 0.0, 0.63}
\definecolor{lasallegreen}{rgb}{0.03, 0.47, 0.19}
\definecolor{hanpurple}{rgb}{0.32, 0.09, 0.98}
\definecolor{green(pigment)}{rgb}{0.0, 0.65, 0.31}

\definecolor{mygray}{gray}{.9}
\usepackage{booktabs}
\newcommand{\topline}{\noalign{\hrule height 0.8 pt}}

\makeatletter
\let\NAT@parse\undefined
\makeatother
\usepackage[pagebackref=false,breaklinks=true,colorlinks,bookmarks=false]{hyperref}
\hypersetup{colorlinks=true,linkcolor={red},citecolor={hanpurple},urlcolor={magenta}}

\title{\LARGE \bf
One-Shot Affordance Grounding of Deformable Objects in Egocentric Organizing Scenes
}

\author{Wanjun Jia$^{1,2}$, Fan Yang$^{1,2}$, Mengfei Duan$^{1,2}$, Xianchi Chen$^{1}$, Yinxi Wang$^{1}$,\\Yiming Jiang$^{1,2}$, Wenrui Chen$^{1,2}$, Kailun Yang$^{1,2,\dag}$, and Zhiyong Li$^{1,2,\dag}$
\thanks{This work was supported in part by the National Key R\&D Program (Grant No. 2022YFB4701400/2022YFB4701404), the National Natural Science Foundation of China (Grant No. U21A20518, No. U23A20341, No. 61976086, and No. 62473139), in part by the Hunan Provincial Research and Development Project (Grant No. 2025QK3019), and in part by the Open Research Project of the State Key Laboratory of Industrial Control Technology, China (Grant No. ICT2025B20).}
\thanks{$^{1}$The authors are with the School of Artificial Intelligence and Robotics, Hunan University, China (email: kailun.yang@hnu.edu.cn, zhiyong.li@hnu.edu.cn).}%
\thanks{$^{2}$The author is also with the National Engineering Research Center of Robot Visual Perception and Control Technology, Hunan University, China.}
\thanks{{\dag}Corresponding authors: Kailun Yang and Zhiyong Li.}%
}

\begin{document}

\maketitle
\thispagestyle{empty}
\pagestyle{empty}

\begin{abstract}
Deformable object manipulation in robotics presents significant challenges due to uncertainties in component properties, diverse configurations, visual interference, and ambiguous prompts. These factors complicate both perception and control tasks. To address these challenges, we propose a novel method for One-Shot Affordance Grounding of Deformable Objects (OS-AGDO) in egocentric organizing scenes, enabling robots to recognize previously unseen deformable objects with varying colors and shapes using minimal samples. Specifically, we first introduce the Deformable Object Semantic Enhancement Module (DefoSEM), which enhances hierarchical understanding of the internal structure and improves the ability to accurately identify local features, even under conditions of weak component information. Next, we propose the ORB-Enhanced Keypoint Fusion Module (OEKFM), which optimizes feature extraction of key components by leveraging geometric constraints and improves adaptability to diversity and visual interference. Additionally, we propose an instance-conditional prompt based on image data and task context, which effectively mitigates the issue of region ambiguity caused by prompt words. To validate these methods, we construct a diverse real-world dataset, AGDDO15, which includes $15$ common types of deformable objects and their associated organizational actions. Experimental results demonstrate that our approach significantly outperforms state-of-the-art methods, achieving improvements of $6.2\%$, $3.2\%$, and $2.9\%$ in KLD, SIM, and NSS metrics, respectively, while exhibiting high generalization performance. Source code and benchmark dataset are made publicly available at \url{https://github.com/Dikay1/OS-AGDO}.

\end{abstract}

\section{Introduction}
Non-rigid/deformable objects, such as clothes, hats, and towels, are crucial manipulation targets for service robots aimed at alleviating daily household chores~\cite{ijcai2024p762,ren2023autonomous}. 
However, compared to rigid objects commonly encountered in robotics, such as mechanical parts and utensils~\cite{li2023locate,yang2024learning}, deformable objects exhibit inherent uncertainty in component information, diverse possible configurations, visual interference, and prompt ambiguity—factors that make them significantly more challenging for autonomous robot perception.

\begin{figure}[!t]
  \centering
\includegraphics[width=0.42\textwidth]{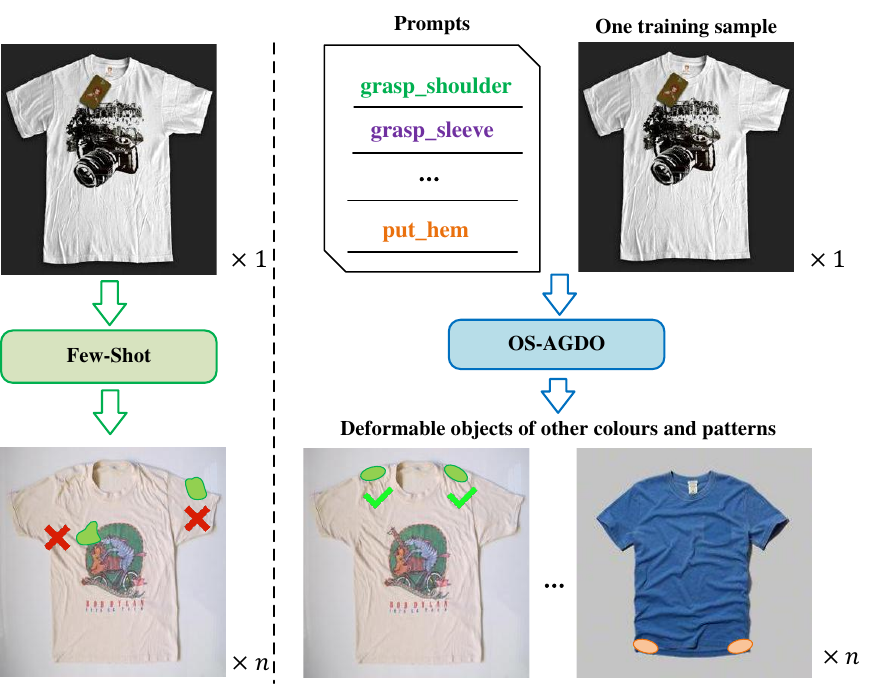}
  \vskip -1ex
  \caption{Comparison of the performance of few-shot method and OS-AGDO on affordance grounding of deformable objects.}
  \label{fig:intro}
  \vskip -4ex
\end{figure}

Current research primarily focuses on manipulation planning and decision-making~\cite{wu2023learning,blanco2023qdp}, with perception modules typically driven by either manually preset rules or data-driven approaches. For instance, Wu~\textit{et al.}~\cite{wu2024unigarmentmanip} manually annotated initial contact points for garment hanging, but their method cannot be generalized to other tasks. The SpeedFolding framework proposed by Avigal~\textit{et al.}~\cite{avigal2022speedfolding} can perform specific folding tasks but requires the collection of tens of thousands of action trajectory data. These approaches have two fundamental limitations: first, feature extraction based on fixed rules cannot adapt to the dynamic feature drift during deformation; second, the large data requirements conflict with the high annotation costs in real-world scenarios. While recent few-shot learning methods~\cite{li2024one,fan2023one} show promise for rigid object manipulation, they prove ineffective for deformable objects due to insufficient component information (\textit{e.g.}, lack of distinct, easily recognizable features and fixed shapes) and the presence of diverse, visually distracting factors (such as varying types, styles, patterns, and colors).
As shown in the left part of Fig.~\ref{fig:intro}, few-shot methods struggle to accurately localize functional regions. They are susceptible to interference from visual factors, such as patterns and shapes, as well as a lack of component information in deformable objects, making local feature extraction difficult.

Inspired by human cognitive mechanisms, we identify three key factors that influence the transfer of experience in manipulating deformable objects: (1) hierarchical understanding of object structure (\textit{e.g.}, decomposing clothes into semantic parts like sleeves, shoulders, etc.); 
(2) decoupled representation of cross-domain features (\textit{e.g.}, separating material textures from geometric deformation features);
(3) verb prompts alone can lead to ambiguity in functional regions (\textit{e.g.}, in garment organization, the verb ``grasp'' could correspond to multiple operational areas, such as the shoulder or sleeves). To address these challenges, we propose the One-Shot Affordance Grounding of Deformable Objects (OS-AGDO) framework, as shown in the right portion of Fig. \ref{fig:intro}. By learning from a single sample of the same category, the robot can recognize multiple previously unseen deformable objects of the same type with varying colors and shapes.
Specifically, we first propose the Deformable Object Semantic Enhancement Module (DefoSEM), which constructs task-relevant attention fields using deformable convolutions. By employing a dual-attention mechanism that combines feature channels and spatial position information, it effectively addresses the problem of hierarchical understanding of object structure. Second, we propose the ORB-Enhanced Keypoint Fusion Module (OEKFM), which extracts keypoints through edge extraction and geometric constraints, improving adaptability to diversity and visual interference.

Finally, we enhance textual prompts by appending specific nouns (\textit{e.g.}, ``shoulder'' or ``sleeve'') after the verb to clarify the scope. By integrating image data and task context, we dynamically generate instance-conditional prompts, which help to resolve ambiguity in functional regions, reduce reliance on fixed templates, and mitigate the risk of overfitting.

Additionally, we have constructed a diverse real-world dataset, AGDDO15, comprising $15$ categories of deformable objects and $15$ affordance types, and performed extensive experiments to validate the effectiveness and superiority of our method. Experimental results show that our approach outperforms existing techniques, achieving improvements of $6.2\%$ in KLD, $3.2\%$ in SIM, and $2.9\%$ in NSS.
To the best of our knowledge, this is the first framework for one-shot affordance learning of deformable objects in egocentric organizing scenes. Our contributions are as follows:
\begin{itemize}
    \item The framework enables the recognition of previously unseen deformable objects with different colors and shapes by learning from a small number of object samples, and achieves effective generalization and adaptive manipulation by learning the ``\textit{Promptable-Object}'' combination only once.
    \item The three key components—DefoSEM, OEKFM, and instance-conditional prompts—address challenges related to uncertainty in component information, diverse configurations, visual interference, and prompt ambiguity in deformable objects, enabling the model to learn diverse features from limited data.
    \item We construct a diverse real-world AGDDO15 dataset, and extensive experiments show the effectiveness and superiority of our proposed approach.
\end{itemize}

\section{Related Work}
\input{Contents/3_RelatedWork}
\section{OS-AGDO: Proposed Framework}
\input{Contents/4_Methodology}

\section{AGDDO15: Established Dataset}
Currently, real-world datasets available for deformable objects from an egocentric perspective are scarce in the field.
To facilitate improved learning of deformable object affordance regions through one-shot learning, we introduce AGDDO15, a dataset designed for the training and evaluation of deformable object affordance grounding, as shown in Fig.~\ref{fig:data}.
Specifically, AGDDO15 consists of $15$ affordance types and $15$ categories of deformable objects.

\subsection{Dataset Collection}
Existing datasets for one-shot affordance learning of deformable objects in organizing scenes are scarce. 
To address this, we created a comprehensive dataset of egocentric images sourced from diverse, high-quality origins to ensure its robustness and completeness.
Portions of the clothes and towel data come from 
DeepFashion ~\cite{liu2016deepfashion} and the FabricFolding Dataset~\cite{he2024fabricfolding}. 
To prevent inadvertently recognizing
deformable objects, such as operators' clothes, during detection in exocentric views, only egocentric images were included to align with manipulation tasks.
Consequently, images from DeepFashion containing human figures or body parts were excluded. 
A random subset of images from the FabricFolding Dataset was chosen for diversity. 
Additional data for other deformable objects, along with extra clothes and towel categories, were sourced from freely licensed websites. 
All deformable objects in the dataset have a coverage rate of over $65\%$.
The final dataset contains $451$ images, with $122$ from existing datasets and $329$ from public sources.

\subsection{Dataset Annotation}
Considering the complexity and typical nature of clothes as deformable objects, we selected seven representative clothes categories. In determining the affordance regions for these objects, we thoroughly examined all potential interaction zones throughout the organizing process. This involved referencing existing garment-folding~\cite{gu2023learning,xue2023unifolding,huang2024rekep,canberk2023cloth,chen2022efficiently,fu2024humanplus,verleysen2020video} and towel-folding~\cite{wu2023learning,he2024fabricfolding,gu2024defnet,weng2022fabricflownet} techniques, as well as hand placement patterns observed in GPT-4o during the organizing scenes of such objects. A total of five participants were involved in the data annotation and organization. Additionally, the entire process was assumed to follow a right-to-left and top-to-bottom sequence, with the ``pick'' action associated with the right side of a symmetrical object and the ``place'' action with the left side. The specific bimanual strategies are shown in Fig.~\ref{fig:process}.

The training set consists of $15$ egocentric images, manually annotated using a sparse annotation method~\cite{fang2018demo2vec}. Interaction regions are defined based on key points identified during the manipulation process, with point density reflecting the frequency of interactions. A Gaussian kernel is applied to each point, generating dense annotations and producing heatmaps of affordance regions. 
Examples of these annotated images are provided in Fig.~\ref{fig:data}~(a).

\subsection{Statistic Analysis}

To gain deeper insights into our AGDDO15 dataset, we highlight its key features from the following aspects.
The category distribution shown in Fig.~\ref{fig:data}~(b), demonstrating the dataset's broad range of affordance/object categories across various scenarios.
The affordance word cloud is presented in Fig.~\ref{fig:data}~(c). The confusion matrix of affordance and object categories is shown in Fig.~\ref{fig:data}~(d). The matrix reveals a multi-to-multi relationship between affordances and deformable object categories, highlighting significant challenges in developing affordance grounding tasks for deformable objects.

\section{Experiments}

\input{Contents/5_Experiments}

\section{Conclusion}
\input{Contents/7_Conclusion}
{\small
\bibliographystyle{IEEEtran}
\bibliography{bib}
}

\end{document}

%% file: Contents/3_RelatedWork.tex
\subsection{Manipulation and Perception of Deformable Objects}
Deformable object manipulation is a critical and complex task in robotics~\cite{fang2020graspnet,kong2023dynamic}. Compared to rigid or articulated objects, deformable objects pose challenges, including uncertain states and diverse patterns and materials, making their manipulation more difficult. 
Current research typically employs reinforcement learning~\cite{lee2021learning,jangir2020dynamic,gu2023learning} or imitation learning approaches~\cite{seita2021learning,lee2024learning}. 
However, precise execution demands accurate perception of object state (\textit{e.g.}, deformation and topology).
Huang~\textit{et al.}~\cite{huang2022mesh} propose a self-supervised loss-based method for occlusion reasoning that reconstructs a cloth mesh and uses a dynamic model for planning.
UniGarmentManip~\cite{wu2024unigarmentmanip} enables robots to generalize to garment manipulation tasks by learning topological visual correspondences across garments with minimal human input, but it still relies on manual key point identification.

Despite advancements, functional region perception for deformable objects remains limited.
Most studies focus on simulation environments, but due to the sim-to-real gap, training with simulation data often fails in real-world settings~\cite{he2024fabricfolding}.
To address these challenges, we propose an affordance grounding method for deformable objects based on real-world environments to solve the gap between simulated and real environments.

\subsection{Affordance Grounding of Deformable Objects}
Affordance grounding for deformable objects focuses on identifying interaction regions. Existing methods typically rely on operational points, including vertex, skeleton, or edge-based localization using object features~\cite{jangir2020dynamic,ge2019deepfashion2}, motion capture systems combined with human demonstration markers, and custom hardware approaches~\cite{yamakawa2011dynamic,shibata2010robotic}. However, current studies predominantly address single-task scenarios, including folding~\cite{avigal2022speedfolding,xue2023unifolding,huang2024rekep}, unfolding~\cite{gu2023learning,ha2022flingbot}, grasping~\cite{chen2023learning,zhang2020learning}, and dressing tasks~\cite{wang2023one,zhang2022learning}.
For instance, Cloth Funnels~\cite{canberk2023cloth} proposes a heuristic folding operation for garments, setting six key points on a long-sleeve T-shirt.
FabricFolding~\cite{he2024fabricfolding} defines four key points for towel manipulation, starting from the top-left corner.
Chen~\textit{et al.}~\cite{chen2022efficiently} compare the frequency of eight grasp points on clothes, finding the back collar is the most frequently used in single-arm fling operations.
ReKep~\cite{huang2024rekep} introduces a vision-based method for optimizing robotic actions, although manual filtering is still required.
Gu~\textit{et al.}~\cite{gu2023learning} apply attention mechanisms to assign higher weights to key points.
Sundaresan~\textit{et al.}~\cite{sundaresan2024learning} compare feature-matching methods for correspondence, including SIFT, SURF, and ORB, and explore correspondence using dense neural networks.

Using heuristic methods, large model dialogue queries, and research into multi-user manipulation habits, we selected $15$ common deformable objects and designed their affordance regions for object organization tasks, resulting in a diverse real-world dataset for deformable object affordance grounding. 
To improve the model's perception of affordance regions, we incorporated an attention mechanism that assigns higher weights to key regions. 
Additionally, geometric constraints were introduced to optimize local region features.

\begin{figure*}
    \centering
    \includegraphics[width=0.88\linewidth]{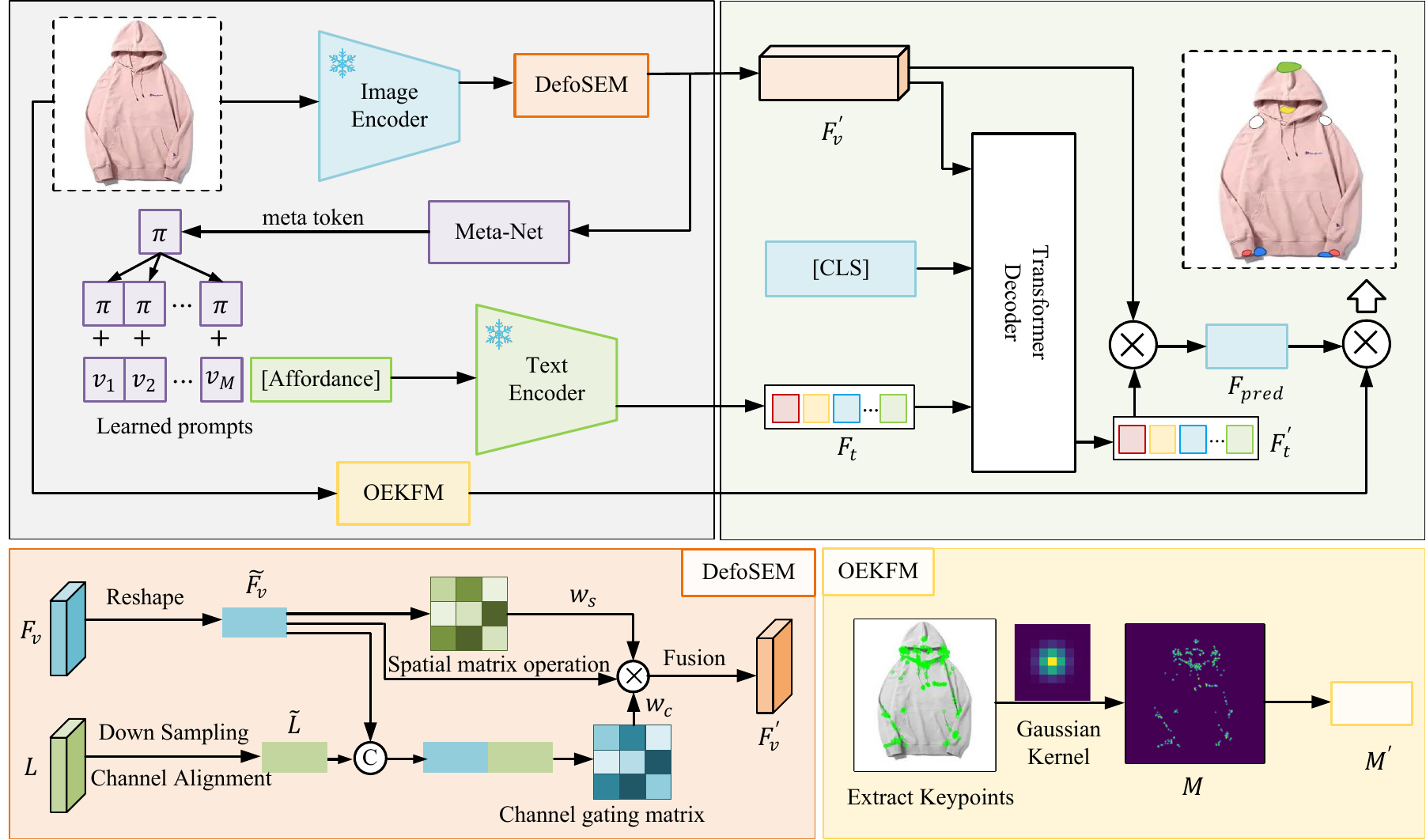}
    \vskip-2ex
    \caption{Illustration of OS-AGDO, the proposed one-shot affordance grounding framework. Our designs are highlighted in four color blocks, which are the visual and text encoders, the CLS-guided transformer decoder, the DefoSEM module, and the Geometric Constraints module. [CLS] denotes the CLS token of the vision encoder.}
    \vspace{-3ex}
    \label{fig:frame}
\end{figure*}

\subsection{One-shot Learning}
Few-shot learning aims to improve model generalization by enabling the recognition of new classes with minimal reference images~\cite{chi2024universal,fu2024humanplus}.
One-shot learning, a subset of few-shot learning, refers to the ability to learn a task or class from a single example or very few samples.
Zhai~\textit{et al.}~\cite{zhai2022one} introduce OSAD-Net, a one-shot manipulation detection method that estimates human action intentions and identifies shared action functions.
This method addresses the challenge of recognizing object manipulation in unseen scenarios.
Ren~\textit{et al.}~\cite{ren2023autonomous} propose a deformable 3D object manipulation framework, enabling robots to generalize to new instances with one demonstration and skill transfer, without retraining.
Li~\textit{et al.}~\cite{li2024one} 
present One-shot Open Affordance Learning (OOAL), which learns object manipulation functions from a single demonstration and enables generalization to new objects with limited data.
Building on these approaches, we propose a one-shot learning method with instance-conditional prompts to improve generalization.

%% file: Contents/4_Methodology.tex
We propose the One-Shot Affordance Grounding of Deformable Objects (OS-AGDO) method, which utilizes a single RGB image and instance-conditional prompts for affordance grounding in egocentric organizing scenes.
During training, affordance labels are used as supervision, with objects divided into \( N \) categories and \( M \) affordance actions. For each category, samples are sequentially input, and the corresponding mask pairs for all \( M \) actions are accessed. After training, the model is evaluated on unseen images to measure its generalization ability. During inference, prior knowledge is used to enhance prediction performance.

The proposed framework, shown in Fig.~\ref{fig:frame}, comprises a visual encoder, a text encoder, and a transformer decoder. The input image \( I_{\text{ego}} \) is passed through the DINOv2~\cite{oquab2023dinov2} visual encoder and the ORB-Enhanced Keypoint Fusion Module (OEKFM). DINOv2 extracts dense patch embeddings, producing visual features \( F_v \in \mathbb{R}^{L \times C} \) and a global [CLS] token. 
The visual features \( F_v \) are passed to the Deformable Object Semantic Enhancement Module (DefoSEM), which we propose to 
enhance functional region features of the target object using dual attention mechanisms, as described in Sec.~\ref{defosem}. 
The enhanced features are sent to both the transformer decoder and the Meta-Net in the CoCoOp module.
The Meta-Net generates dynamic prompts, which, combined with affordance labels, are processed by the CLIP text encoder to produce text embeddings \( F_t \in \mathbb{R}^{N \times C} \), as explained in Sec.~\ref{cocoop}. 
Simultaneously, we propose OEKFM, which extracts keypoints and generates geometric feature regions using a Gaussian kernel, as detailed in Sec.~\ref{oekfm}. 
The transformer decoder receives the visual features \( F_{v}^{'} \), text embeddings \( F_t \), and the [CLS] token, producing an initial affordance prediction. This prediction \( F_{\text{pred}} \) is subsequently optimized using the geometric features from OEKFM.

\begin{figure*}[!t]
  \centering
\includegraphics[width=0.85\textwidth]{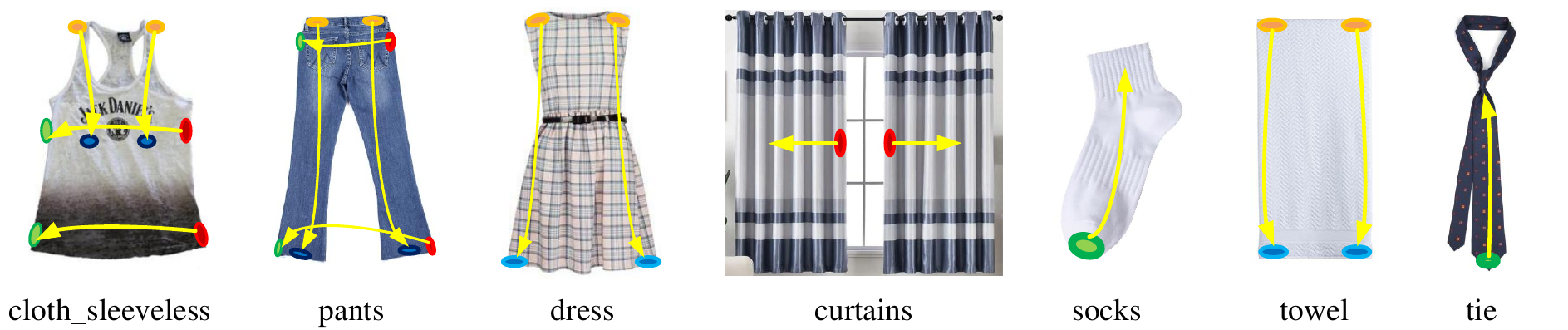}
  \vskip -2ex
  \caption{Strategies of OS-AGDO for organizing various categories of deformable objects.}
  \label{fig:process}
  \vskip -3.5ex
\end{figure*}

\subsection{DefoSEM}
\label{defosem}
As shown in the orange box in Fig.~\ref{fig:frame}, we enhance the visual features extracted by DINOv2. The DefoSEM combines image label \( L \) with visual features \( F_v \), improving feature learning through a multi-scale gating mechanism.

First, adaptive average pooling is applied to downsample the image labels, reducing the spatial dimensions while preserving global information. Next, a convolutional operation is used to align the channel dimensions of the label features with the visual features. The visual features are then adjusted to generate \( \tilde{F_v} \), and spatial gating weights \( W_s \) are computed. These weights modulate the visual features along the spatial dimension, emphasizing important spatial regions.
Simultaneously, the adjusted visual features \( \tilde{F_v} \) and the dimension-processed label features \( \tilde{L} \) are concatenated along the channel dimension. A convolution operation then generates channel gating weights \( W_c \), which are used to weigh the visual features and highlight important channel-specific information.
Finally, the adjusted visual features \( \tilde{F_v} \), channel gating weights \( W_c \), and spatial gating weights \( W_s \) are fused to generate the enhanced features \( F_{v}^{'}  \). 
The fusion formula is as follows:
\vspace{-0.8ex}
\begin{equation}
F_{v}^{'}  = \tilde{F_v}~ \times (0.5 + W_c) \times (0.5 + W_s).
\end{equation}
\vspace{-3ex}

This fusion strategy stabilizes the features by enhancing both channel and spatial information, resulting in feature maps with richer semantic information.

\subsection{ORB-Enhanced Keypoint Fusion Module}
\label{oekfm}
As shown in the yellow box in Fig.~\ref{fig:frame}, we introduce a new branch after inputting the image \( I_{\text{ego}} \), referred to as OEKFM. First, we apply feature matching using the ORB algorithm~\cite{rublee2011orb} to extract keypoints of the deformable object. These keypoints are then processed using a Gaussian kernel to generate a feature region \( M \).

To enhance the affordance region and filter out irrelevant background or incorrectly identified regions, we weight the transformer decoder's prediction \( {F_\text{pred}} \) by the adjusted keypoint feature region \(M^{\prime}\), yielding the final prediction \( P \).
\begin{equation}
P = F_{\text{pred}}\odot M^{\prime}.
\end{equation}

To ensure that both \( M \) and  \( P \) lie within the range \([0, 1]\), \( M \) is normalized. Additionally, to enhance its effect, the value range of \( M^{\prime} \) is adjusted to \([1, 2]\):
\begin{equation}
M^{\prime} = 1 + \frac{M - M_{\text{min}}}{M_{\text{max}} - M_{\text{min}} + \epsilon}.
\end{equation}

The small constant \( \epsilon {=} 1e{-}8 \) is introduced to avoid division by zero errors. Finally, to ensure the proper backpropagation of loss, the weighted prediction results \( P \) are clamped to the range \([0, 1]\) using the following clamp function:
    \begin{equation}
    P_{\text{final}} = \text{clamp}(P, 0, 1).
    \end{equation}
    
The definition of the clamp function is as follows:
    \begin{equation}
    \text{clamp}(x, 0, 1) = 
    \begin{cases} 
    0, & \text{if } x < 0, \\
    x, & \text{if } 0 \leq x \leq 1, \\
    1, & \text{if } x > 1. 
    \end{cases}
    \end{equation}

This function guarantees that \( P_{\text{final}} \) always stays within the desired range of \([0, 1]\).
By incorporating this module, we better achieve cross-domain generalization of deformable objects' variability and complexity, enhancing the recognition of human-object interaction regions and accurately filtering out inaccurate or erroneous localization areas.

\begin{figure*}
    \centering
    \includegraphics[width=\linewidth]{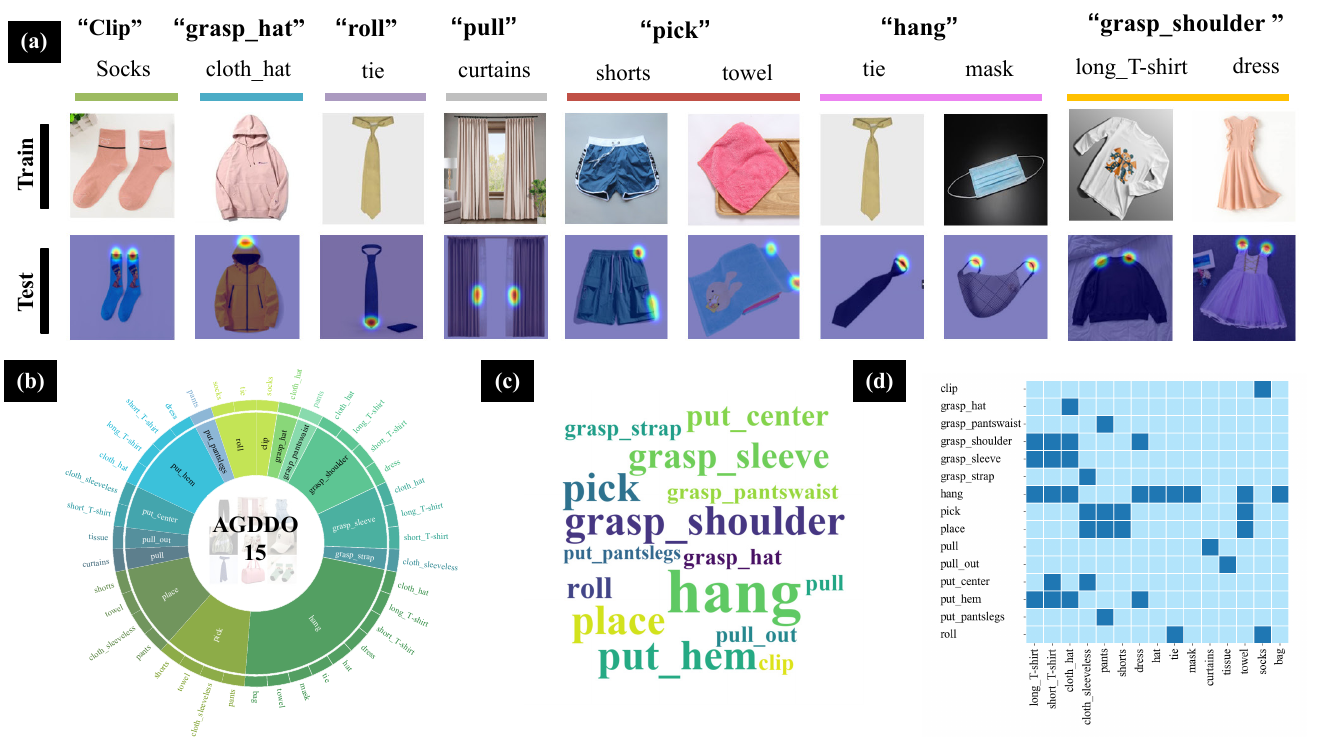}
    \vskip -3ex
    \caption{\textbf{Properties of the established AGDDO15 dataset.}
    \textbf{(a)} Some examples from the dataset. 
    \textbf{(b)} The distribution of categories in AGDDO15.
    \textbf{(c)} Illustrates the affordance word cloud.
    \textbf{(d)} Confusion matrix between the affordance category and the object category in AGDDO15, where the horizontal axis denotes the object category and the vertical axis denotes the affordance category.}
    \vspace{-4ex}
    \label{fig:data}
\end{figure*}

\subsection{Instance-conditional Prompt}
\label{cocoop}
As illustrated in the purple module of Fig.~\ref{fig:frame}, \(v_m\) represents the learnable context vector for the $m$-th input (where \(m {\in} \{1, 2, \dots, M\}\)), \(x\) denotes the feature vector of the image, and \(\pi\) represents the instance-conditional generated by Meta-Net.
By incorporating the CoCoOp method~\cite{zhou2022conditional}, \(N\) neural networks are constructed to generate \(N\) context tokens. 
Specifically, a lightweight neural network, called Meta-Net, is learned on top of the context vector. 
This network generates a conditional token for each input, which is combined with the task-specific affordance labels.
The vector $v_m$ is a static learnable parameter, and $v_m(x)$ denotes its instance-conditioned version:
\(v_m(x) {=} v_m {+} \pi\). 
The initialized learnable context vectors \(\{v_1(x), v_2(x), \dots, v_M(x)\}\) are inserted before the text CLS token, allowing for shared functionality across all. 
Thus, the text prompt is: 
\(t_i(x) {=} \{v_1(x), v_2(x), \dots, v_M(x), c_i \}\). Here $g(\cdot)$ is the text encoder.
The final prediction is calculated as:
\vspace{-1.2ex}
\begin{equation}
p(y|x)=\frac{exp(sim(x,g(t_y(x)))/\tau)}{\sum_{i=1}^{K}exp(sim(x,g(t_y(x)))/\tau)}.
\end{equation}
\vspace{-2.3ex}

During training, we update the context vectors \(\left \{v_m\right \} _{m=1}^{M}\) alongside the parameters of Meta-Net. In this work, Meta-Net is constructed using a two-layer bottleneck architecture (Linear-ReLU-Linear), where the hidden layer reduces the input dimensions by a factor of $24$.
The integration of this module enhances the semantic clarity and task relevance of the prompts, helping the model better understand task requirements and accurately locate correct functional regions.

%% file: Contents/5_Experiments.tex
\subsection{Implementation Details}

\begin{figure*}
    \centering
    \includegraphics[width=0.98\linewidth]{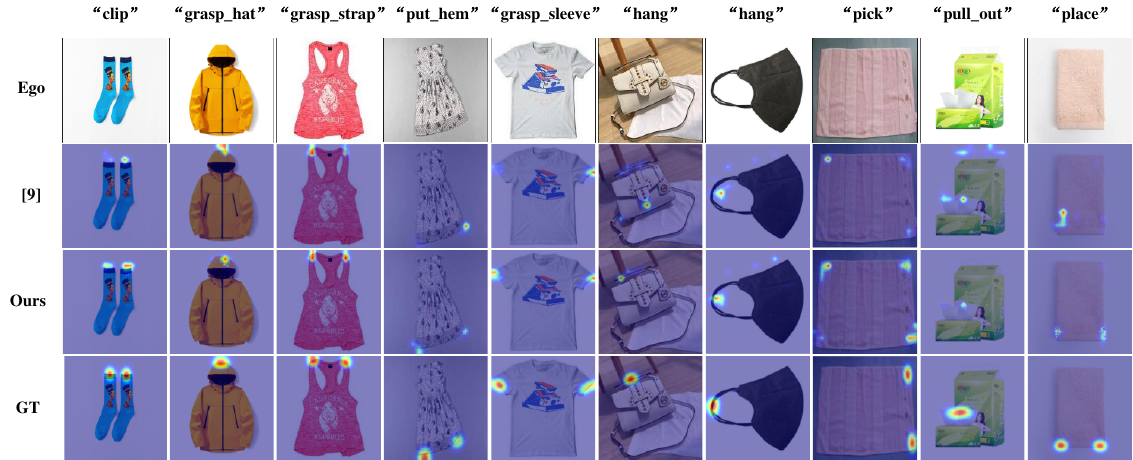} 
    \vskip-2ex
    \caption{\textbf{Visual affordance grounding predictions on the AGDDO15 dataset.} Qualitative comparison of our method with the state-of-the-art functionality grounding method (OOAL~\cite{li2024one}).}
    \vspace{-4ex}
    \label{fig:viz}
\end{figure*}

Experiments were conducted on an RTX 3090 GPU. All visual models were based on the same Vision Transformer (ViT-base) architecture. 
The training was performed for $10k$ iterations, with each iteration using a self-centered image as input. Initially, the image was resized to $256{\times}256$, followed by a random crop to $224{\times}224$ pixels, with random horizontal flipping applied. 
The model was trained using the SGD optimizer with a learning rate of $0.01$.

To evaluate the results on the AGDDO15 dataset, we used commonly employed metrics, including Kullback-Leibler Divergence (KLD), Similarity (SIM), and Normalized Scanpath Saliency (NSS), as in previous works~\cite{li2024one}.

\subsection{Results of Functional Affordance Grounding}
We present the state-of-the-art method for one-shot affordance grounding and its performance.
Under the same experimental setup, as shown in Table~\ref{table:comparion}, our method significantly outperforms competing approaches across all metrics. Specifically, compared to OOAL~\cite{li2024one}, our method achieves improvements of $6.2\%$, $3.2\%$, and $2.9\%$ in KLD, SIM, and NSS, respectively.
These gains stem from our method’s tailored design for deformable object affordance grounding.
Our approach more effectively handles the diversity and complexity of deformable objects in affordance grounding by extracting more accurate functional region features and enhancing the model’s generalization ability, leading to superior performance across the evaluated metrics.

We present grounding visualizations of the baseline method, our method, and the GT, as shown in Fig.~\ref{fig:viz}. 
Compared to the baseline, our approach integrates dual attention and geometric constraints to filter out irrelevant background regions, enhancing fine-grained localization of functional areas. 
Our approach significantly reduces background interference (see the 2nd and 9th columns of Fig.~\ref{fig:viz}), focusing affordance predictions on the key regions of the target object.
Additionally, it effectively addresses challenges posed by the placement of flexible objects at varying angles by adaptively correcting rotational distortions. 
As illustrated in the 4th and 6th columns of Fig.~\ref{fig:viz}, 
our method rectifies localization errors where the baseline misidentifies or overlooks regions, demonstrating superior robustness and adaptability.
Furthermore, to improve accuracy, we enhance the extraction of contour edge information, particularly for functional regions often found along object edges.
As seen in the 7th and 8th columns of Fig.~\ref{fig:viz}, our method precisely aligns affordance regions with the object edges, making it more suitable for real-world applications.

\begin{table}[!t]
    \centering
    \caption{Comparison to SOTA methods on the AGDDO15 dataset for relevant tasks (↑/↓ indicates higher/lower is better)}
    \vskip-2ex
    \begin{tabular}{lccc}
        \topline
        \rowcolor{mygray}
        Model & KLD $\downarrow$ & SIM $\uparrow$ & NSS $\uparrow$ \\
        \hline\hline
        OOAL~\cite{li2024one}  & 0.913 & 0.532 & 2.174 \\
        Ours  & 0.856 & 0.549 & 2.236 \\
        \bottomrule
    \end{tabular}
    \label{table:comparion}
	\vspace{-2.5ex}
\end{table}

\begin{table}[!t]
    \centering
    \caption{Ablation results of proposed modules. The \textbf{best} is highlighted in bold.}
    \vskip-1ex
    \begin{tabular}{l c c c}
    \topline
    \rowcolor{mygray}
    Model    & KLD $\downarrow$ & SIM $\uparrow$ & NSS $\uparrow$ \\ 
    \hline\hline
    Baseline & 0.913                     & 0.532                     & 2.174                    \\
    +DefoSEM & 0.896                     & 0.551                     & 2.191                    \\
    +CoCoOp & 0.890                      & 0.545                     & 2.200                    \\ 
    +OEKFM   & \textbf{0.856}            & \textbf{0.549}            & \textbf{2.236}           \\ 
    \bottomrule
    \end{tabular}
    \label{table:ablation}
	\vspace{-4.5ex}
\end{table}

\subsection{Ablation Study}

\vspace{3pt}\textbf{Impact of each design.}
We use OOAL~\cite{li2024one} as the baseline and progressively integrate our method to analyze the impact of each proposed design. The results in Table~\ref{table:ablation} demonstrate that each module consistently leads to significant performance improvements. 
Specifically, the addition of the CoCoOp module effectively mitigates overfitting and significantly enhances performance on the KLD and NSS metrics. However, the gain in the SIM metric is less pronounced. This is because the CoCoOp module tends to learn task-relevant features rather than strictly matching the overall distribution, and the dynamically generated prompts may introduce some noise, limiting their effect on the SIM metric. 
In contrast, the inclusion of the geometric constraint module improves robustness to multi-angle and multi-scale variations, resulting in further improvements across all metrics.

\vspace{3pt}\textbf{Number of keypoint features.}
As shown in Table~\ref{table:n_feature}, we evaluate the effect of different feature quantities, \( n {=} \{200, 400, 600, 800, 1000\} \), on the KLD, SIM, and NSS metrics. The results indicate that the optimal performance across all metrics is achieved when \( n {=} 400 \). 
This finding aligns with our algorithm design principle: an excessively high number of features may cause the model to focus on background regions or other irrelevant areas, while an insufficient number of features can result in inadequate extraction of critical information, such as edge details.

\vspace{3pt}\textbf{Components of DefoSEM.}
From the information in Table~\ref{table:defosem}, it can be observed that using only channel attention enables the model to effectively capture the importance of different channels. However, it overlooks critical spatial information, resulting in imprecise affordance region localization, particularly at the edges and fine details of deformable objects. Conversely, employing only spatial attention allows the model to focus on salient spatial regions but fails to distinguish the contributions of different feature channels. This limitation reduces its adaptability to complex textures and multi-scale variations of deformable objects, ultimately affecting its overall perception of functional regions. 
When both channel and spatial attention mechanisms are integrated, the model can simultaneously capture important inter-channel features and critical spatial information, significantly improving the accuracy of affordance grounding.

\begin{table}[!t]
    \centering
    \caption{Impact of varying $n$ on KLD, SIM, and NSS metrics.}
    \vskip-2ex
    \begin{tabular}{lc c c}
    \topline
    \rowcolor{mygray}
    $n$ & KLD $\downarrow$ & SIM $\uparrow$ & NSS $\uparrow$ \\
    \hline\hline
    1000         & 0.894          & 0.535          & 2.205          \\
    800          & 0.885          & 0.547          & 2.211             \\
    600          & 0.954          & 0.523          & 2.110     \\
    \textbf{400} & \textbf{0.856} & \textbf{0.549} & \textbf{2.236} \\
    200          & 0.895          & 0.546          & 2.191    \\
    \bottomrule
    \end{tabular}
    \label{table:n_feature}
	\vspace{-2ex}
\end{table}

\begin{table}[!t]
    \centering
    \caption{Components of DefoSEM. `Channel' denotes channel attention. `Spatial' denotes spatial attention.}
    \vskip-2ex
    \begin{tabular}{c|c|ccc}
    \topline
    \rowcolor{mygray}
    Channel & Spatial & KLD $\downarrow$ & SIM $\uparrow$ & NSS $\uparrow$ \\
    \hline\hline
    \ding{51}    &                 & 0.984                 & 0.511                 & 2.095                 \\
                 & \ding{51}       & 0.907                 & 0.548                 & 2.193                 \\
    \textbf{\ding{51}}    & \textbf{\ding{51}}      & \textbf{0.856}        & \textbf{0.549}        & \textbf{2.236}        \\
    \bottomrule
    \end{tabular}
    \label{table:defosem}
	\vspace{-4.5ex}
\end{table}

\subsection{Performance on Real Deformable Objects}

\begin{figure}[!t]
  \centering
\includegraphics[width=0.48\textwidth]{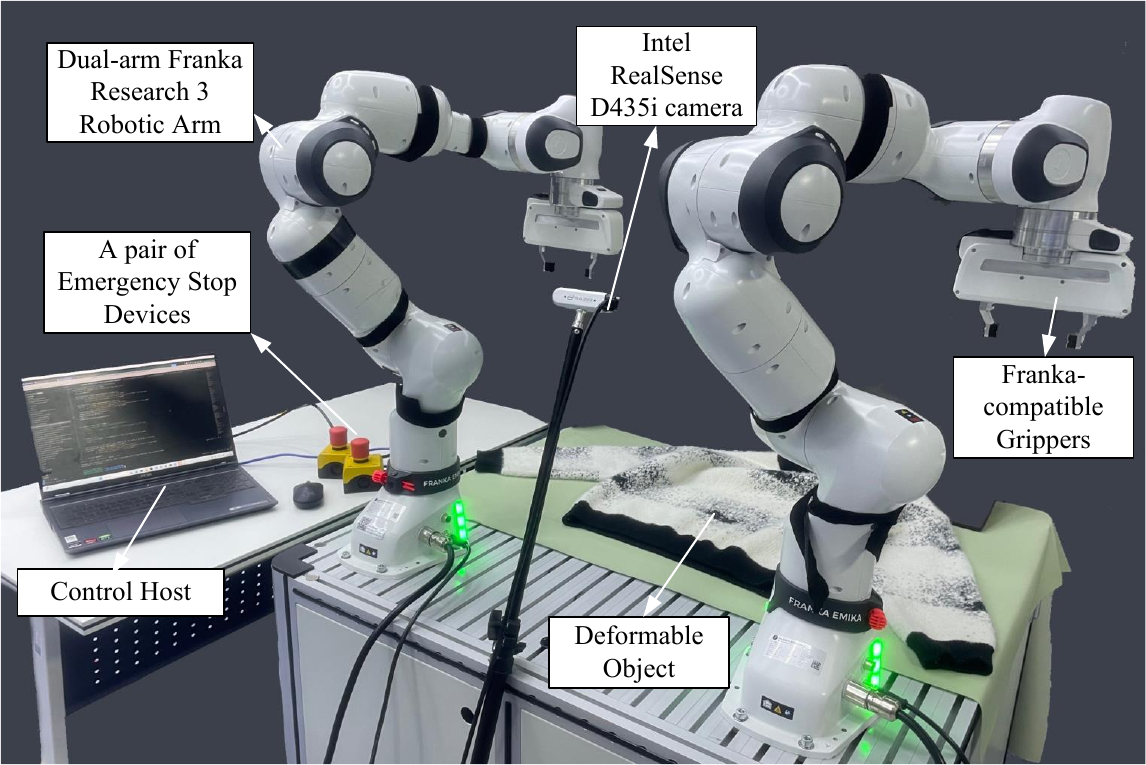}
  \vskip -1.5ex
  \caption{Real-world experimental platform for deformable objects.}
  \label{fig:platform}
  \vskip -2ex
\end{figure}

To provide insights into practical deployment, we set up a real-world experimental platform, as shown in Fig.~\ref{fig:platform}. 
This platform consists of a dual-arm Franka Research 3 robotic arm, two Franka-compatible grippers, an Intel RealSense D435i camera, a pair of emergency stop devices, a deformable object (a piece of clothing), and a control computer. The model was deployed on the control computer, which received visual inputs from the camera and communicated with the robotic arm via a network interface to perform affordance grounding and execute the experimental tasks.

\begin{figure}[!t]
  \centering
\includegraphics[width=0.48\textwidth]{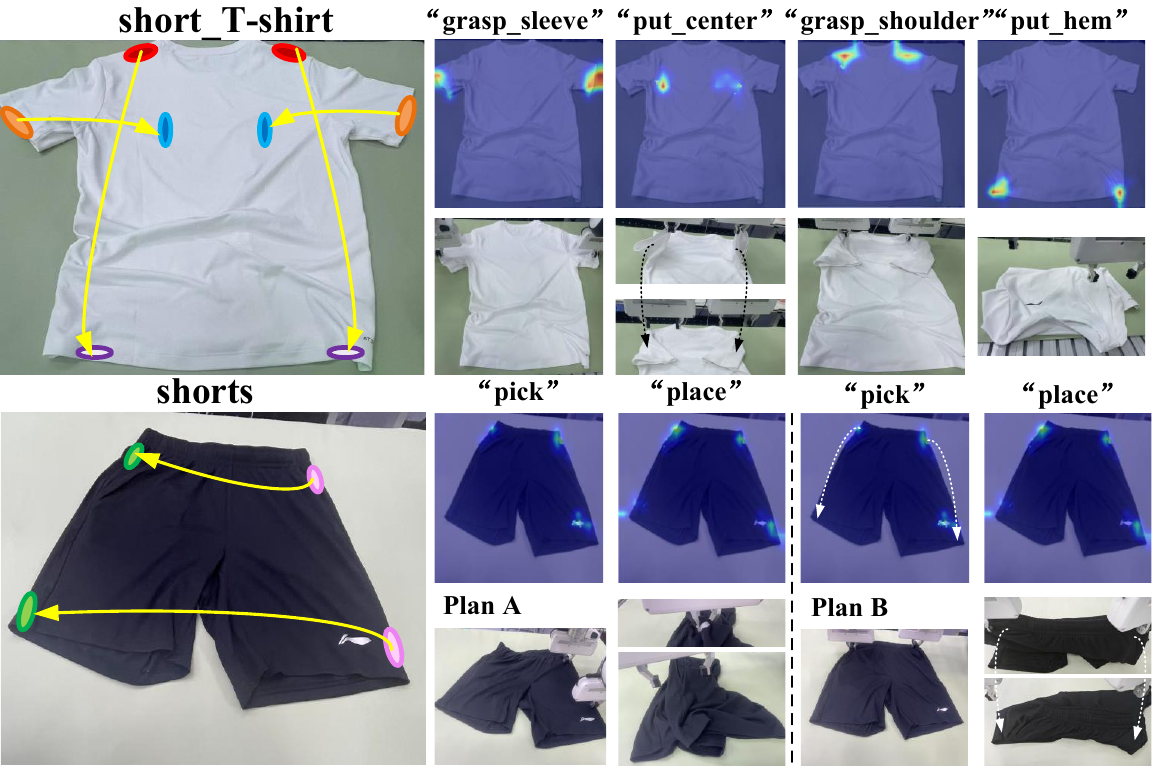}
  \vskip -1.5ex
  \caption{Examples of deformable object affordance grounding in organizing scenes based on our OS-AGDO.}
  \label{fig:experiment}
  \vskip -3.5ex
\end{figure}

As shown in Fig.~\ref{fig:experiment}, we present the results of our method on deformable objects. The left panel illustrates the affordance-based organizing strategy where the model sequentially identifies predefined functional regions based on prompts.
For instance, when organizing a short-sleeve T-shirt, the model first identifies the ``grasp\_sleeve'' region, followed by the ``put\_center'' region, then ``grasp\_shoulder,'' and finally ``put\_hem.'' The first row displays a visualization of our method's affordance grounding. 
After identifying all affordance regions, we proceed with the robotic arm operation experiments.
In the experiment involving shorts, our annotation design specified ``pick'' actions in the two right-side regions and ``place'' actions in the two left-side regions (see ``Plan A''). However, after learning from the top-to-bottom annotations in the towel and pants categories, a new strategy emerged for shorts, following a top-to-bottom pattern in which ``pick'' actions occurred at the two waist regions, and ``place'' actions were performed at the two pant leg regions. Notably, in the experiment, this strategy successfully completed the organization task (see ``Plan B''). These results demonstrate that our method effectively facilitates affordance grounding for deformable objects and validates the practical utility of the model’s learned strategy.

%% file: Contents/7_Conclusion.tex
This paper proposes a novel method for One-Shot Affordance Grounding of Deformable Objects (OS-AGDO) in egocentric organizing scenes, enabling the recognition of functional regions in deformable objects with unseen colors and shapes from a few samples.
We propose DefoSEM to improve semantic relationships and enhance multi-region perception accuracy. 
Geometric constraints are introduced through OEKFM to capture robust edge points, enhancing localization precision for deformable objects.
Additionally, we resolve ambiguity in functional regions by appending specific nouns to verb-only prompts and dynamically generating instance-conditional prompts using both image data and task context.
To facilitate this research, we have constructed AGDDO15, a diverse real-world dataset for training and evaluation of affordance grounding of deformable objects.
Experiments show our method significantly outperforms state-of-the-art approaches, achieving more accurate localization of potential interaction regions, as well as generalization from basic to novel categories and prompt transfer across datasets. 
Our work offers new insights into efficient functional region localization and robust manipulation of deformable objects in real-world scenarios.
In the future, we intend to tackle affordance grounding of deformable objects in open environments with multimodal fusion and large vision models.